\documentclass[]{aidata}

\usepackage[toc,page,header]{appendix}
\usepackage{minitoc}
\usepackage{cleveref} 
\usepackage{subcaption}
\usepackage{booktabs}
\usepackage[table]{xcolor}
\definecolor{mycolor}{RGB}{240,248,255} % AliceBlue
\usepackage[table,dvipsnames]{xcolor}
\usepackage{array}
\usepackage{tabularx}
\usepackage{ragged2e}
\usepackage[ruled,vlined,linesnumbered]{algorithm2e}
\usepackage{multirow}

\definecolor{myblue}{RGB}{232, 240, 254}
\definecolor{mypurple}{RGB}{243, 229, 245}
\definecolor{mygreen}{RGB}{232, 245, 233}
\definecolor{lightblue}{RGB}{230,240,255}
\definecolor{lightpink}{RGB}{255,230,240}
\definecolor{lightyellow}{RGB}{255,250,230}
\definecolor{lightgreen}{RGB}{230,255,230}
\definecolor{lightpurple}{RGB}{240,230,255}
\definecolor{reflectbg}{RGB}{245,245,250}
\definecolor{repibg}{RGB}{245,250,245}
\definecolor{qualifybg}{RGB}{255,250,240}
\definecolor{cvprblue}{RGB}{0,0,139}
\definecolor{mycolor}{RGB}{240,248,255}
\definecolor{PineGreen}{RGB}{0,128,0}
\definecolor{OrangeRed}{RGB}{255,69,0}
\definecolor{tkcolor}{RGB}{93, 163, 232}

\newcolumntype{C}{>{\Centering\arraybackslash}X}

\usepackage{amsmath,amssymb,mathtools,amsthm}

\theoremstyle{plain}

\theoremstyle{definition}

\theoremstyle{remark}

\usepackage[most]{tcolorbox}

\newtcbox{\hlprimarytab}{on line, rounded corners, box align=base,
  colback=green!10, colframe=white, size=fbox, arc=3pt,
  before upper=\strut, top=-2pt, bottom=-4pt, left=-2pt, right=-2pt, boxrule=0pt}

\newtcbox{\hlsecondarytab}{on line, rounded corners, box align=base,
  colback=red!10, colframe=white, size=fbox, arc=3pt,
  before upper=\strut, top=-2pt, bottom=-4pt, left=-2pt, right=-2pt, boxrule=0pt}

\newtcolorbox{takeaways}[2][]{
    width=\columnwidth,
    toprule=0.0pt,
    leftrule=0.9pt,
    bottomrule=0.9pt,
    rightrule=0.9pt,
    arc=0pt,
    colback = tkcolor!10, 
    colframe = tkcolor, 
    boxsep=0pt,left=7pt,right=7pt,top=4pt,bottom=4pt,
    fontupper=\linespread{0.1}\selectfont,
    title=#2,#1,
    before skip=0.7em,
    after skip=0.7em
}

\usepackage{listings}
\tcbuselibrary{skins, breakable}

\usepackage{natbib}
\usepackage{latexsym}
\usepackage{url}
\usepackage{pifont}
\usepackage{makecell}
\usepackage{xspace}
\usepackage{adjustbox}
\usepackage{hyperref}
\usepackage[edges]{forest}
\usepackage{tikz}
\usepackage{caption}
\usepackage{amsfonts}
\usepackage{float}
\usepackage{stfloats}
\usepackage{colortbl}

\hypersetup{
    colorlinks,
    linkcolor={blue!80!black},
    citecolor={blue!80!black},
}

\title{Credit Where It's Due: Cross-Modality Connectivity Drives Precise Reinforcement Learning for MLLM Reasoning}

\author[1,2,4,*]{Zhengbo Jiao}
\author[2,*]{Shaobo Wang}
\author[3,5]{Zifan Zhang}          % ← affiliation corrected to [1,4]
\author[1]{Wei Wang}
\author[1,\textsuperscript{$\blacktriangle$}]{Bing Zhao}  % ← Project Leader with black triangle
\author[1,\dagger]{Hu Wei}
\author[2,\dagger]{Linfeng Zhang}

\affiliation[1]{AI DATA, Alibaba Group Holding Limited}
\affiliation[2]{EPIC Lab, SJTU}
\affiliation[3]{Skylenage}
\affiliation[4]{SUFE}
\affiliation[5]{WHU}

\contribution[*]{Equal contribution}

\contribution[\textsuperscript{$\blacktriangle$}]{Project leader}
\contribution[\dagger]{Corresponding authors}

\abstract{

Reinforcement Learning with Verifiable Rewards (RLVR) has significantly advanced the reasoning capabilities of Multimodal Large Language Models (MLLMs), yet how visual evidence is integrated during reasoning remains poorly understood. We explore multimodal RLVR through the lens of cross-modal attention connectivity and find that only a small fraction of tokens (approximately 15\%) exhibit strong visual-textual coupling. These high-connectivity tokens act as anchors that ground reasoning in the image, while the majority follow linguistic patterns. During RLVR training, credit assignment naturally concentrates on these anchors, sharpening their visual grounding over time. Building on this insight, we propose Anchor-Token Reinforcement Learning (AT-RL), a lightweight framework that selectively reinforces high-connectivity tokens via graph-based clustering of attention topology. Evaluated across the Qwen2.5-VL series (3B–32B), AT-RL introduces only 1.2\% overhead yet enables the 32B model to surpass the 72B-Instruct baseline on MathVista (80.2), with consistent gains observed across STEM, vedio and general tasks. Conversely, training solely on low-connectivity tokens causes severe degradation, confirming that effective multimodal RL hinges on precise credit assignment to visual anchors. Our work reveals that reasoning quality is governed not by token quantity but by the fidelity of cross-modal anchoring.
}

\vspace{-12pt}

\renewcommand{\date}[1]{}          % 覆盖原定义
\renewcommand{\correspondence}[1]{} % 覆盖原定义

\renewcommand{\checkdatalist}{}    % 彻底移除所有 checkdata 内容

\usepackage{etoolbox}
\makeatletter
\patchcmd{\mymaketitle}
  {\ifdefempty{\checkdatalist}{\vspace*{0.65cm}}{\checkdatalist\par}}
  {\ifdefempty{\checkdatalist}{}{\checkdatalist\par}} % 空时不再插入空间
  {}{}
\makeatother

\begin{document}

\maketitle
\vspace{-20pt}
\begin{figure}[H]
    \centering
    \includegraphics[width=1\linewidth]{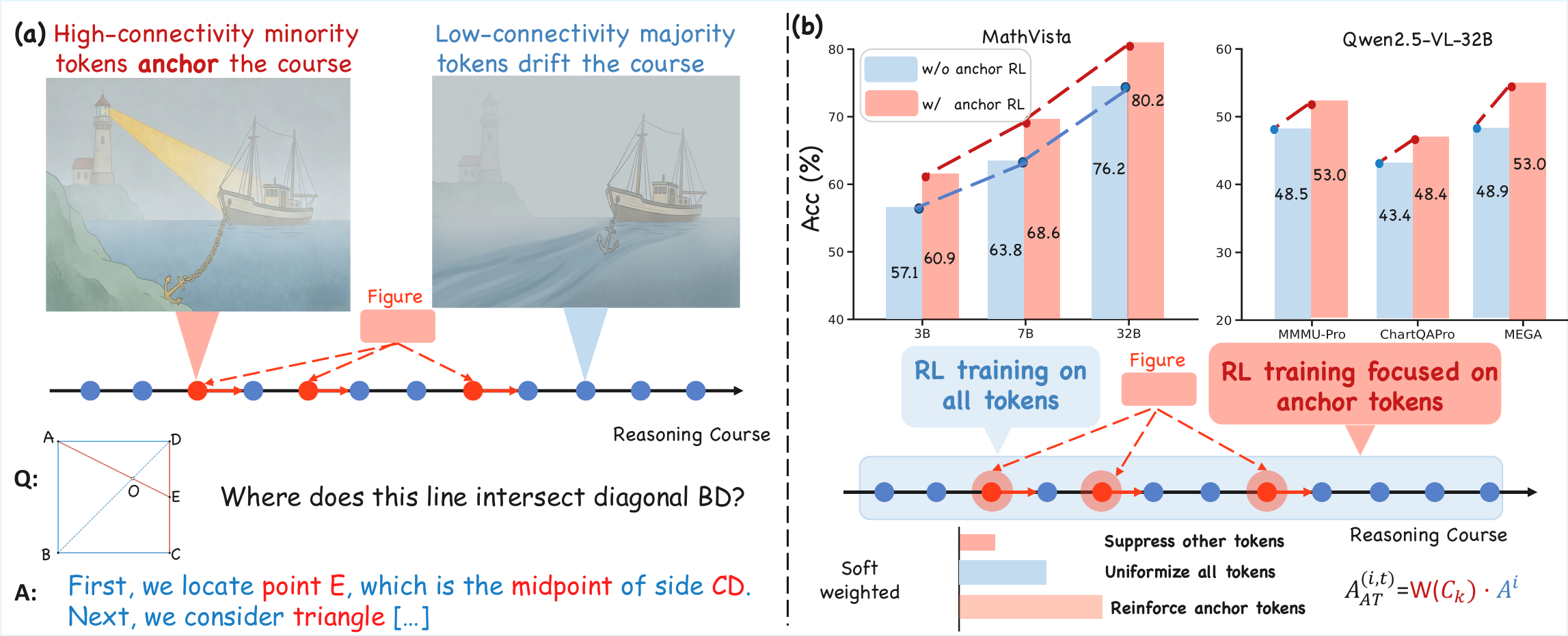}
   \caption{(a) In multimodal CoTs, only a minority of tokens (approximately 15\%) exhibit high cross-modal connectivity and act as ``perceptual anchors'' that ground reasoning in visual evidence, while the majority are low-connectivity tokens fulfilling linguistic structures. (b) RLVR using AT-RL to modulate reinforcement signals via cluster-based soft weighting yields substantial reasoning improvements that scale from 3B to 32B. By precisely optimizing these critical anchors, our 32B model achieves 80.2 on MathVista and 56.6 on MathVerse. }

    \label{fig:placeholder}
\end{figure}

\section{Introduction}
\label{sec:intro}
Multimodal Large Language Models (MLLMs) have demonstrated significant proficiency in complex domains such as mathematics and programming. 
While pre-training provides foundational capabilities, post-training through Reinforcement Learning (RL) has become the primary methodology for eliciting structured Chain-of-Thought (CoT) reasoning. 
The success of models such as OpenAI o1 and DeepSeek-R1 highlights the effectiveness of RL in optimizing reasoning trajectories, proving that models can resolve sophisticated problems by exploring an expanded solution space. 
Central to this development is Reinforcement Learning with Verifiable Rewards (RLVR), where online algorithms such as GRPO, GSPO, SAPO, and DAPO serve as scalable frameworks for multimodal policy optimization. 
In these reasoning tasks, the model processes a heterogeneous input sequence composed of interleaved visual patches and linguistic tokens, yet it generates a response serialized as a purely textual sequence. 
Consequently, while multimodal RLVR has achieved impressive gains through outcome-based rewards, existing implementations indiscriminately broadcast credit across all tokens, failing to address the fundamental challenge of distinguishing which tokens genuinely incorporate visual evidence versus those that merely follow linguistic patterns.

In this paper, we analyze the underlying mechanisms of multimodal RLVR through an innovative lens of attention connectivity patterns, investigating how the connectivity, representing the aggregated attention weights between textual tokens and visual patches, impacts reasoning performance. 
We first point out that in the Chain-of-Thought (CoT) processes of MLLMs, the connectivity distribution exhibits a distinct pattern where the majority of tokens are generated with low connectivity, while a critical minority of tokens emerge with high connectivity. 
Through comparing the functional roles of these two groups, we observe that the tokens with high connectivity, which we define as \textit{perceptual anchors}, function as the lighthouse that grounds the reasoning course in visual evidence, while the tokens with lower connectivity act as the engine that completes linguistic structures and maintains computational coherence, as depicted in Figure 1(a). 
In addition to qualitative analysis, we conduct controlled experiments by modulating the reinforcement weights based on connectivity levels during the policy gradient update. 
Quantitative results reveal that a soft weighting approach, which prioritizes high-connectivity anchors while retaining the contribution of other tokens, leads to substantial improvements in reasoning performance across diverse benchmarks, whereas uniform weighting or hard truncation yields suboptimal gains, confirming the importance of a connectivity-based credit assignment. 

Furthermore, by analyzing the evolution of connectivity patterns during training, we find that our proposed AT-RL framework enables the reasoning model to undergo a structured optimization of its attention distribution. 
Unlike standard RLVR, which often inherits the base model topology without significant refinement, AT-RL progressively sharpens the connectivity of perceptual anchors to improve grounding precision, thereby more effectively distinguishing them from the linguistic-heavy tokens. 
The above observations highlight the critical role that connectivity-based credit assignment may play in bridging visual inputs and textual logic during training, provided that the reinforcement signal is properly guided.

Building upon the discovery of perceptual anchors, we propose \textit{Anchor-Token Reinforcement Learning} (AT-RL), a lightweight and plug-and-play framework for connectivity-driven advantage assignment. 
Specifically, AT-RL constructs a token-level graph from aggregated attention weights and employs graph partitioning to group reasoning sequences into semantically coherent clusters. 
For each cluster, we derive a weight based on its inter-modal connectivity density to re-scale the sequence-level advantage into a fine-grained, token-level credit signal. 
Unlike binary gradient masking in methods like VPPO that risks losing essential reasoning context, our soft-weighting strategy concentrates learning signals on perceptual anchors while preserving other tokens with modulated weights to maintain linguistic and computational coherence. 
Extensive experiments across the Qwen2.5-VL series (3B to 32B) show that AT-RL consistently outperforms strong baselines including GRPO, DAPO, and VPPO. 
Notably, our 32B model trained with AT-RL surpasses the performance of the significantly larger 72B-Instruct model on multiple benchmarks. 
Furthermore, AT-RL exhibits robust generalization across diverse tasks including multimodal STEM reasoning, general VQA, and complex video reasoning. 
Conversely, we find that random weighting, reverse weighting, or training exclusively on other tokens leads to severe performance degradation, confirming that multimodal RLVR efficacy is primarily driven by the precise optimization of cross-modal anchors.

Finally, we conduct detailed ablation studies to investigate the impact of key hyperparameters and the underlying mechanics of AT-RL. 
We find that our connectivity-driven weighting formula naturally concentrates reinforcement credit on approximately 15\% of the tokens representing the perceptual anchors, while dynamically scaling the advantage of all other tokens based on their semantic clusters. 
This continuous weighting scheme ensures that the learning signal is precisely focused on visual-linguistic alignment without compromising the linguistic coherence of the reasoning chain. 
Notably, a single set of hyperparameters including clustering scales and connectivity thresholds generalizes across all evaluated benchmarks ranging from STEM reasoning to VideoMMMU, which underscores the intrinsic stability of the framework. 
Profiling results further show that the entire AT-RL module is remarkably efficient, introducing a negligible computational overhead of approximately 1.2\% per training iteration. 
These findings confirm that the efficacy of multimodal RLVR is primarily driven by the precise cluster-based modulation of cross-modal anchors, a strategy that aligns reinforcement signals with the structural interdependencies of vision and language to achieve reasoning precision. In summary, our main contributions are as follows:
\begin{itemize}
    \item \textbf{Discovery of Perceptual Anchors:} We identify a structural dichotomy in the attention topology of multimodal Chain-of-Thought reasoning, where a critical minority of tokens act as \textit{perceptual anchors} that ground the reasoning process in visual evidence, while the remaining tokens fulfill linguistic roles.
    \item \textbf{The AT-RL Framework:} We propose \textit{Anchor-Token Reinforcement Learning}, a lightweight, plug-and-play framework that constructs a token-level graph from attention weights and employs graph partitioning to group sequences into semantically coherent clusters. By re-scaling sequence-level advantages into fine-grained credit signals based on inter-modal connectivity density, AT-RL enables precise optimization while preserving essential linguistic context.
    \item \textbf{Scalable Generalization and Efficiency:} We demonstrate that AT-RL exhibits robust generalization across diverse tasks---including multimodal STEM reasoning, general VQA, and complex video reasoning---using a single set of hyperparameters. Our results show consistent gains across model scales from 3B to 32B, where Qwen2.5-VL-32B-Instruct model trained with ours notably surpasses the Qwen2.5-VL-72B-Instruct model with a negligible computational overhead of only 1.2\%.
\end{itemize}

\section{Related Work}

\textbf{Multimodal reasoning bottlenecks.} Recent work systematically reveals that insufficient visual perception capability constitutes the core bottleneck limiting multimodal reasoning performance. MATH-Vision first evaluates multimodal mathematical reasoning on authentic competition problems and finds that models significantly underperform in vision-requiring tasks compared to pure textual scenarios \citep{zhang2024mathvision}. MathVerse further confirms that most models fail to genuinely comprehend visual information in diagrams and instead rely on textual cues for reasoning \citep{liu2024mathverse}. ErrorRadar demonstrates through error detection tasks that models struggle to identify inconsistencies between visual inputs and logical reasoning in their own outputs \citep{yang2024errorradar}. We-Math 2.0 reveals that verbose chain-of-thought responses often mask deficient understanding of fundamental visual principles \citep{liu2025wemath}. GeoPQA quantifies a pronounced geometric perception gap, showing that humans achieve near-perfect structural recognition while multimodal models frequently fail despite strong linguistic capabilities \citep{peng2025geopqa}. BLINK-Twice distinguishes that models lack the fine-grained visual analysis required for complex reasoning \citep{kumar2025blink}. VisualToolBench quantifies that 70 to 80 percent of failures originate from visual perception errors rather than reasoning deficits \citep{guo2025beyond}. MMErroR further validates that models cannot accurately localize error sources within visual-linguistic contexts \citep{lam2026mmerror}. Collectively these studies indicate that current multimodal models frequently generate textual reasoning chains without authentic grounding in visual inputs, establishing visual perception deficiency as the primary constraint on multimodal reasoning capability.

\textbf{Reinforcement learning for LLM.} Reinforcement learning for large language models originated with PPO as the standard optimizer for reinforcement learning from human feedback (RLHF) \citep{ppo}. DPO later reframed alignment as a supervised learning problem, bypassing explicit reinforcement learning \citep{dpo}. The emergence of reasoning-capable models catalyzed Reinforcement Learning with Verifiable Rewards (RLVR), where GRPO established the foundation through group-relative advantage estimation without explicit value networks \citep{deepseekmath}. DAPO introduced dynamic sampling and clip-higher mechanisms to mitigate gradient dilution \citep{DAPO}. GVPO enhanced training convergence via group variance control \citep{gvpo}. TreeRPO extended policy optimization to tree-structured reasoning trajectories \citep{treerpo}. GSPO enforced sequence-level policy updates for improved trajectory coherence \citep{gspo}. GMPO adopted geometric-mean objectives for stable policy distribution optimization \citep{gmpo}. GTPO resolved gradient conflicts in multi-step reasoning through trajectory-based advantage estimation \citep{gtpo}. Lambda-GRPO reinterpreted GRPO as an implicit process reward model \citep{lambda_grpo}. CE-GPPO controlled policy entropy via gradient-preserving clipping \citep{ce_gppo}. MAPO proposed a mixed advantage framework for balanced exploration \citep{mapo}. SAPO introduced soft adaptive mechanisms to dynamically regulate policy update intensity \citep{gao2025softadaptivepolicyoptimization}. Despite algorithmic diversity, these methods uniformly broadcast credit across all tokens, overlooking the heterogeneous roles tokens play in visual grounding.

\textbf{Fine-Grained Credit Assignment.} Recent research increasingly recognizes that different tokens play heterogeneous roles during reasoning. TokenDPO and Critical Tokens were the first to validate the feasibility of token-level optimization in preference alignment tasks \citep{tokendpo,critical_tokens}. Beyond the 80/20 Rule further reveals that only a few high-entropy tokens can drive performance improvement in pure text reasoning \citep{high_entropy_tokens}. GCPO extends this idea to image generation tasks, improving generation quality by reinforcing structural key tokens \citep{gcpo_image}. In multimodal scenarios, StepGRPO introduces step-level rewards for fine-grained supervision, though its effectiveness relies on high-quality chain-of-thought for cold-start \citep{stepgrpo}. VPPO proposes the concept of token perception, reinforcing visually-dependent tokens through binary gradient masking \citep{vppo}. Recent works further explore domain-specific fine-grained optimization: Zoom-Zero targets video temporal localization, SketchVL focuses on chart understanding, and Video-KTR selectively reinforces tokens by combining multiple modality-aware signals for video reasoning \citep{zoomzero,sketchvl,videoktr}. These methods collectively demonstrate that aligning credit assignment with the specific functional roles of tokens is an effective direction for enhancing model capabilities.
\section{Preliminaries}
\subsection{Cross-Modal Attention and Debiasing}
\label{subsec:cross-modal-attention-debiasing}

The cross-modal attention weight from a generated textual token at index \(i\) to an input visual token at index \(j\) in layer \(l\) and head \(h\) is defined as:
\[
A^l_h[i,j] = \frac{\exp(e_{ij})}{\sum_{k}\exp(e_{ik})}, \quad e_{ij} = \frac{(W^Q \mathbf{o}_i) \cdot (W^K \mathbf{o}_j)^\top}{\sqrt{d}}.
\]
Here, \(\mathbf{o}_i\) and \(\mathbf{o}_j\) are the hidden states at positions \(i\) and \(j\), \(W^Q\) and \(W^K\) are learnable projection matrices, and \(d\) is the dimensionality of the attention head. This computation follows the standard attention mechanism \cite{vaswani2017attention}. When \(i\) indexes a textual token and \(j\) a visual token, \(A^l_h[i,j]\) quantifies the model's visual grounding at step \(i\), representing the raw, uncalibrated dependency of the generated text on a visual patch.

\textbf{``Attention bias'' corresponds to systematic distortions in the attention distribution.} The raw attention is contaminated by positional preferences and inflated scores for padding tokens, i.e., \(A_{\text{observed}} = A_{\text{semantic}} + A_{\text{bias}}\). We introduce a parametric calibration function \(\mathcal{B}\) to remove this bias, obtaining a purified weight \(\tilde{A}^l_h[i,j] = \mathcal{B}(A^l_h[i,j]; \Theta)\) for fine-grained credit assignment, extending recent debiasing work \cite{zhao2025attention}.

\subsection{Reinforcement Learning with Verifiable Rewards}
\textbf{Reinforcement Learning with Verifiable Rewards (RLVR)} optimizes policies through deterministic, rule-based feedback. In reasoning tasks such as mathematics, the correctness of a model output $y$ can be verified against the ground-truth $y_{\text{gt}}$ using a binary reward function:
\begin{equation}
    R(y, y_{\text{gt}}) =
    \begin{cases}
        +1, & \text{if } y = y_{\text{gt}}, \\
        -1, & \text{otherwise}.
    \end{cases}
\end{equation}

\textbf{Proximal Policy Optimization (PPO)} \citep{ppo} serves as the foundational policy-gradient algorithm for this paradigm. It stabilizes updates by constraining the probability ratio $r_t(\theta) = \pi_{\theta}(a_t|s_t) / \pi_{\theta_{\text{old}}}(a_t|s_t)$ within a clipped range $[1-\epsilon, 1+\epsilon]$. The objective is formulated as:
\begin{equation}
    \mathcal{L}_{\mathrm{PPO}}(\theta) = \mathbb{E}_t \left[ \min\left( r_t(\theta) A_t,~ \text{clip}(r_t(\theta), 1-\epsilon, 1+\epsilon) A_t \right) \right]
\end{equation}
where $A_t$ is the advantage typically estimated by an additional value network.

\textbf{Group Relative Policy Optimization (GRPO)} \citep{deepseekmath} adapts the PPO framework to the RLVR setting by substituting the value network with group-normalized rewards. For a group of $G$ responses $\{o_1, \dots, o_G\}$ generated from a prompt $q$, the sequence-level advantage $A^{(i)}$ is computed as:
\begin{equation}
A^{(i)} = \frac{R_i - \mathrm{mean}(\{R_j\}_{j=1}^G)}{\mathrm{std}(\{R_j\}_{j=1}^G)}
\end{equation}
The final objective applies the PPO clipped surrogate loss sequence-wise and incorporates a KL term:
\begin{equation}
\mathcal{L}_{\mathrm{GRPO}}(\theta) = \frac{1}{G} \sum_{i=1}^G \frac{1}{|o_i|} \sum_{t=1}^{|o_i|} \min\big(r_t(\theta)A^{(i)}, \text{clip}(r_t(\theta), 1-\epsilon, 1+\epsilon)A^{(i)}\big) - \beta D_{\mathrm{KL}}(\pi_{\theta} \,\|\, \pi_{\mathrm{ref}})
\end{equation}
Subsequent advancements, including DAPO \citep{DAPO}, GSPO \citep{gspo}, and SAPO \citep{gao2025softadaptivepolicyoptimization}, build upon this framework to further enhance training stability and convergence. Our proposed AT-RL leverages these engines as a base.

\section{Anchor-Token Reinforcement Learning (AT-RL)}
\begin{figure}[tb!]
    \centering
    \includegraphics[width=\linewidth]{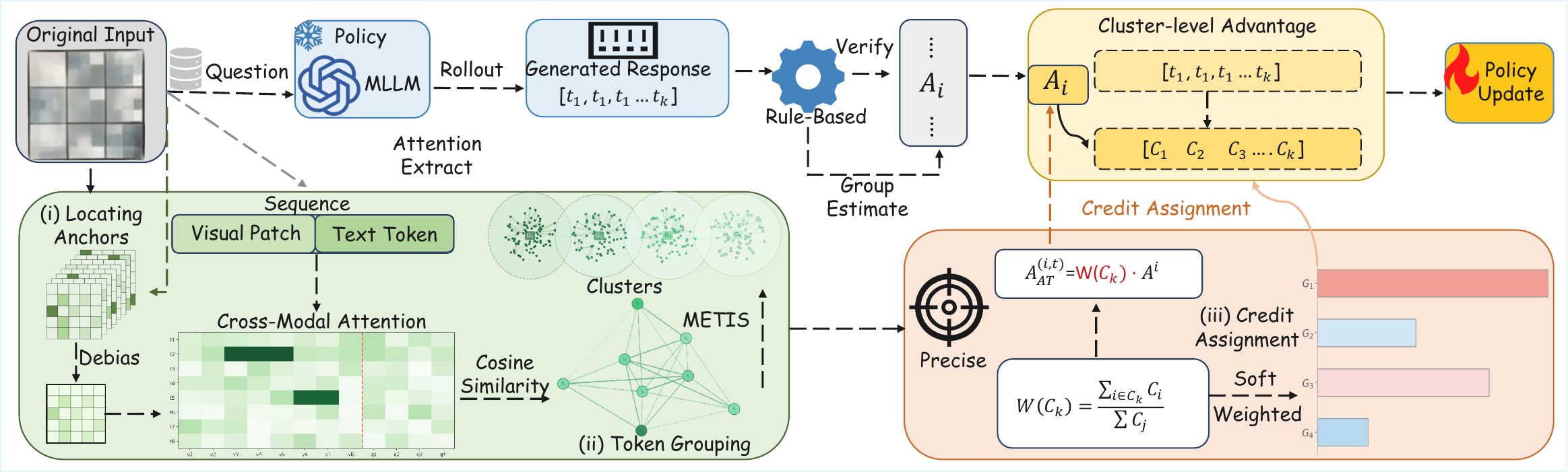}
    \caption{\textbf{Overview of the Anchor-Token Reinforcement Learning (AT-RL) framework.} Our method refines the reinforcement signal through a three-stage connectivity-aware analysis: 
    \textbf{(i) Locating Anchors:} Extracting cross-modal attention footprints from the MLLM and applying debiasing to identify key perceptual tokens; 
    \textbf{(ii) Token Grouping:} Constructing a functional dependency graph based on attention similarity and partitioning it into semantic clusters $\{C_k\}$ via the METIS algorithm; 
    \textbf{(iii) Credit Assignment:} Quantifying cluster-level importance $W(C_k)$ to perform soft-weighting on the base advantage $A^i$. 
    The resulting modulated advantage $A_{AT}^{(i,t)}$ ensures that policy gradients are prioritized for tokens essential to visual grounding while preserving the logical coherence of the reasoning chain.}
    \label{fig:pipeline}
\end{figure}

We introduce \textbf{AT-RL} (\textbf{A}nchor-\textbf{T}oken \textbf{R}einforcement \textbf{L}earning), a policy optimization framework that resolves the uniform-credit limitation of existing RLVR algorithms. By leveraging intrinsic cross-modal attention, AT-RL concentrates reinforcement signals on critical tokens that ground reasoning in visual evidence, ensuring that gradients are prioritized for tokens essential to task success. As illustrated in Fig.~\ref{fig:pipeline}, our method unfolds in three stages: (i) locating perceptual anchors via connectivity quantification, (ii) grouping tokens into coherent clusters through graph-based partitioning, and (iii) reinforcing credit assignment of advantage signal.

\subsection{Locating Perceptual Anchors}

The first stage of AT-RL is to extract and calibrate internal signals from the Multimodal Large Language Model (MLLM) to identify the functional roles of tokens. Given a generated reasoning sequence $S = \{t_1, \dots, t_T\}$ and the input visual patches $\mathcal{V}$, we extract the self-attention matrices $\mathbf{A}^l_h \in \mathbb{R}^{T \times |S|}$ from the top $L$ transformer layers and $H$ attention heads. To obtain a stable representation of the model's semantic footprint, we aggregate these matrices across both layers and heads:
\begin{equation}
    \bar{\mathbf{A}} = \frac{1}{L H} \sum_{l=1}^L \sum_{h=1}^H \mathbf{A}^l_h.
\end{equation}

\noindent\textbf{Bias Correction.}
As discussed in Section 3.1, raw attention weights are often contaminated by structural artifacts, such as ``attention sinks'' at initial tokens or positional preferences, which do not reflect genuine semantic grounding. To isolate meaningful signals, we apply the debiasing operator $\mathcal{B}(\cdot)$ by normalizing each column $j$ of the aggregated matrix with a mean-normalized bias curve $b_j$:
\begin{equation}
    \tilde{\mathbf{A}}_{i,j} = \frac{\bar{\mathbf{A}}_{i,j}}{b_j}.
\end{equation}
The curve $b_j$ is parameterized by a combination of exponential decay and periodic terms to capture standard positional biases (see Appendix~\ref{sec:impl} for the exact formulation). This yields the calibrated attention matrix $\tilde{\mathbf{A}}$ used for subsequent credit assignment.

\noindent\textbf{Quantifying Connectivity Density.}
Based on the calibrated matrix $\tilde{\mathbf{A}}$, we define the \textbf{Connectivity Density} $C_i$ for each generated textual token $t_i$. This metric quantifies the intensity with which a token extracts visual evidence during generation:
\begin{equation}
    C_i = \sum_{j \in \mathcal{V}} \tilde{\mathbf{A}}_{i,j}.
\end{equation}
Empirical observations across various reasoning tasks reveal a distinct structural dichotomy in multimodal traces. A critical minority of tokens—the \textbf{Perceptual Anchors}—exhibit disproportionately high connectivity density, typically corresponding to key observations or evidence extraction from the image. In contrast, the vast majority of tokens, which we refer to as \textbf{Other tokens}, possess $C_i$ values near zero, as they primarily serve to maintain linguistic coherence and logical transitions. This heterogeneity suggests that the uniform-credit assumption is physically inaccurate, providing the foundation for our non-uniform credit assignment.

\subsection{Graph-based Token Grouping}

While connectivity density $C_i$ provides a per-token measure of grounding, individual token signals can be noisy or fragmented. To identify coherent reasoning steps, we group tokens into semantic clusters by representing the sequence as a graph and applying topological partitioning.

\noindent\textbf{Graph Construction.}
We construct a weighted undirected graph $G = (V, E)$, where each vertex $v_i \in V$ corresponds to a token $t_i$ in the reasoning sequence. The edge weight $w_{ij}$ between two tokens $t_i$ and $t_j$ is defined by the cosine similarity of their calibrated attention footprints (rows of $\tilde{\mathbf{A}}$):
\begin{equation}
    w_{ij} = \frac{\tilde{\mathbf{A}}_i \cdot \tilde{\mathbf{A}}_j}{\|\tilde{\mathbf{A}}_i\| \|\tilde{\mathbf{A}}_j\|}.
\end{equation}
To ensure a sparse and semantically relevant graph, an edge is established only if $w_{ij} > \tau_{\mathrm{sim}}$, where $\tau_{\mathrm{sim}}$ is a similarity threshold. This graph captures the functional dependencies within the reasoning trace, where tokens with similar visual grounding and linguistic contexts are more densely connected.

\noindent\textbf{Sequence Partitioning via METIS.}
We partition the graph $G$ into $K$ disjoint semantic clusters $\{C_1, \dots, C_K\}$ using the METIS algorithm \cite{karypis1998fast}. METIS is chosen for its efficiency in minimizing the edge cut—the total weight of edges crossing cluster boundaries—thereby ensuring that tokens within the same cluster share maximum semantic affinity. The number of clusters $K$ is set proportional to the sequence length $T$ to adapt to varying reasoning complexities.

\noindent\textbf{Structural Refinement and Propagation.} 
To ensure the robustness of the identified clusters, we perform a two-step refinement process to handle attention noise and semantic continuity. First, we apply \textit{centroid-based denoising} to suppress spurious connections that may bypass initial partitioning. For each cluster $C_k$, we compute its semantic centroid $\mathbf{c}_k$ in the attention space. Tokens that exhibit low cosine similarity to their respective centroid ($\cos(\tilde{\mathbf{A}}_i, \mathbf{c}_k) < \tau_{\mathrm{cen}}$) are attenuated by a factor $\alpha$, thereby purifying the cluster’s signal. Second, to preserve the continuity of reasoning steps that transcend discrete cluster boundaries, we execute a \textit{topological propagation} step. We identify central nodes with high out-degree within the top-$q$ percentile and propagate their reinforcement influence to their $R$-nearest neighbors. This mechanism effectively recovers transition tokens (Other tokens) that are topologically vital to the Perceptual Anchors, resulting in a set of semantically dense clusters that serve as the foundational units for fine-grained credit assignment.

\subsection{Reinforced Credit Assignment}

The final stage of AT-RL translates the topological structures identified in Section 4.2 into differential reinforcement signals. The core of this mechanism lies in modulating the advantage signal at the cluster level, thereby resolving the limitations of uniform credit assignment in multimodal reasoning.

\noindent\textbf{Quantifying Cluster Importance.}
To measure the contribution of each reasoning step to visual perception, we define the cluster weight $W(C_k)$. This weight is calculated as the proportion of the sequence's total connectivity density that is concentrated within a specific cluster $C_k$:
\begin{equation}
    W(C_k) = \frac{\sum_{i \in C_k} C_i}{\sum_{j=1}^T C_j}.
\end{equation}
This quantitative metric reflects the ``perceptual load'' across different segments of the reasoning trace. Clusters containing Perceptual Anchors naturally receive higher weights, while those primarily serving linguistic transitions are assigned lower weights.

\noindent\textbf{Modulated Advantage and Objective.}
AT-RL does not alter the fundamental loss structure of the underlying RL engine; instead, it acts as an advantage modulation layer. As illustrated in Fig. 1(b), given the original sequence-level advantage $A^{(i)}$ generated by a base engine (e.g., GRPO), we map it to a token-level modulated advantage $A_{\text{AT}}^{(i,t)}$:
\begin{equation}
    A_{\text{AT}}^{(i,t)} = W(C_k) \cdot A^{(i)}, \quad \forall t \in C_k.
\end{equation}
By reshaping the advantage signal based on semantic roles, we integrate this modulated signal into the standard policy gradient objective. The final AT-RL optimization function is formulated as follows:
\begin{equation}
    \mathcal{L}_{\text{AT-RL}}(\theta) = \frac{1}{G} \sum_{i=1}^G \frac{1}{|o_i|} \sum_{t=1}^{|o_i|} \min \left( r_t(\theta) A_{\text{AT}}^{(i,t)}, \text{clip}(r_t(\theta), 1-\epsilon, 1+\epsilon) A_{\text{AT}}^{(i,t)} \right) - \beta D_{\mathrm{KL}}(\pi_{\theta} \,\|\, \pi_{\mathrm{ref}}).
\end{equation}

By substituting the uniform $A^{(i)}$ with the fine-grained $A_{\text{AT}}^{(i,t)}$, AT-RL ensures that policy gradients are prioritized for tokens essential to task success. This soft-weighting scheme reinforces visual grounding while preserving the linguistic integrity of the reasoning chain. Unlike binary masking, this approach maintains low-weight gradients for non-anchor tokens, ensuring that the model does not lose its ability to generate logical and coherent transitions. Because the modulation occurs at the advantage level, AT-RL is inherently plug-and-play, allowing for seamless integration into existing engines like PPO, GRPO, or SAPO without requiring modifications to the value network or the policy architecture.

\section{Experiments}
\label{sec:experiments}

\subsection{Experimental Setup}
\label{subsec:setup}
\noindent \textbf{Models.}  
We conduct training experiments on Qwen2.5-VL-3B-Instruct~\citep{qwen25vl_3b,qwen25vl_blog}, Qwen2.5-VL-7B-Instruct~\citep{qwen25vl_7b,qwen25vl_blog}, and Qwen2.5-VL-32B-Instruct~\citep{qwen25vl_32b,qwen25vl_blog}. For zero-shot comparison, we additionally evaluate the Qwen2.5-VL-72B-Instruct variant~\citep{qwen25vl_72b,qwen25vl_blog} alongside proprietary models including Llama 4 Scout~\citep{meta2025llama4}, Claude 3.5 Sonnet~\citep{anthropic2024claude35,aisi2025claude35}, Gemini 2.0 Flash~\citep{google2025gemini20}, and GPT-4o~\citep{openai2024gpt4o}.

\noindent \textbf{Training Data.}  
Main RL experiments are trained on the \textbf{ViRL-39K} dataset~\citep{virl39k} for one epoch. Cross-dataset generalization studies employ Geometry-3K~\citep{lu2021intergps} for 7B models and GeoQA-8K~\citep{chen2022geoqa} for 3B models.

\noindent \textbf{Benchmarks.}  
Evaluation spans five categories: (1) \textit{Mathematical reasoning}: GeoQA$_{\text{test}}$, MathVerse~\citep{liu2024mathverse}, MathVision~\citep{zhang2024mathvision}, MathVista~\citep{lu2024mathvista}, WeMath~\citep{liu2025wemath}; (2) \textit{Multi-discipline}: MMMU-Pro~\citep{yue2025mmmcupro}, MMMU\_Val~\citep{yue2024mmmu}, EMMA\_core~\citep{emma2025}; (3) \textit{Real-world}: MEGA~\citep{megabench2024}; (4) \textit{Chart understanding}: ChartQAPro~\citep{masry2025chartqapro}, ChartMimic~\citep{yang2024chartmimic}; (5) \textit{Video reasoning}: Video-R1~\citep{videor12025} (including VSI-Bench~\citep{vsibench2024}, VideoMMMU~\citep{videommmu2025}, MMVU~\citep{zhao2025mmvu} subsets).

\noindent \textbf{Evaluation Protocol.}  
We follow the standard evaluation protocol from the LLMs-Eval benchmark suite~\citep{llm-eval}. All methods are evaluated on \textit{testmini} splits using Acc@1 at temperature 0.1. Video experiments test frame inputs of 16, 32, and 64 frames. WeMath reports overall accuracy only.

\noindent \textbf{Baselines.}  
We compare against representative RLVR methods: (1) \textbf{GRPO}~\citep{deepseekmath,grpo}: group-relative policy optimization with intra-group reward normalization; (2) \textbf{DAPO}~\citep{DAPO}: integrates clip-higher, dynamic sampling, token-level loss, and overlong reward shaping; (3) \textbf{GSPO}~\citep{gspo}: sequence-likelihood-based policy optimization with sequence-level clipping; (4) \textbf{SAPO}~\citep{gao2025softadaptivepolicyoptimization}: replaces hard clipping with smooth, temperature-controlled soft gating and uses asymmetric temperatures for positive/negative advantages to stabilize MoE training; (5) \textbf{StepGRPO}~\citep{stepgrpo}: step-level credit assignment via explicit process supervision; (6) \textbf{FT-RL}~\citep{high_entropy_tokens}: forks token-level RL on high-entropy decision points; (7) \textbf{VPPO}~\citep{vppo}: leverages token perception by reweighting trajectory advantage based on visual dependency and focusing updates on perceptually pivotal tokens.

\noindent \textbf{Infrastructure.}  
All training runs use 8×NVIDIA A100 GPUs with identical hyperparameters across methods. Complete hyperparameter settings are provided in the supplementary material.

\subsection{Main Results}
\label{subsec:main_results}

\begin{table}[ht]
\centering
\caption{
Main results of Qwen2.5-VL-7B-Instruct with different RLVR training methods, evaluated on the \textit{testmini} split of five multi-modal reasoning benchmarks and their average (Avg). 
Arrow values represent absolute percentage-point changes relative to the \emph{Zero-shot baseline}.
{\scriptsize\textcolor{ForestGreen}{$\uparrow$}} indicates improvement and {\scriptsize\textcolor{red}{$\downarrow$}} indicates decline.
The best results are highlighted in \textbf{bold}.
}
\label{tab:main_results_7b}
\resizebox{\textwidth}{!}{%
\begin{tabular}{l|ccccc|c}
\toprule
\textbf{Method} & GeoQA$_{\text{test}}$ & MathVerse & MathVision & MathVista & WeMath & \textbf{Average} \\
\midrule
Zero-shot & 44.05 & 39.62 & 22.70 & 61.10 & 57.59 & 45.01 \\
\midrule
\rowcolor{gray!10} \multicolumn{7}{l}{\textit{Uniform Credit Baselines}} \\
GRPO & 46.86{\scriptsize\textcolor{ForestGreen}{$\uparrow$2.81}} & 41.81{\scriptsize\textcolor{ForestGreen}{$\uparrow$2.19}} & 24.37{\scriptsize\textcolor{ForestGreen}{$\uparrow$1.67}} & 63.80{\scriptsize\textcolor{ForestGreen}{$\uparrow$2.70}} & 59.71{\scriptsize\textcolor{ForestGreen}{$\uparrow$2.12}} & 47.31{\scriptsize\textcolor{ForestGreen}{$\uparrow$2.30}} \\
DAPO & 47.54{\scriptsize\textcolor{ForestGreen}{$\uparrow$3.49}} & 42.61{\scriptsize\textcolor{ForestGreen}{$\uparrow$2.99}} & 25.36{\scriptsize\textcolor{ForestGreen}{$\uparrow$2.66}} & 64.30{\scriptsize\textcolor{ForestGreen}{$\uparrow$3.20}} & 60.41{\scriptsize\textcolor{ForestGreen}{$\uparrow$2.82}} & 48.04{\scriptsize\textcolor{ForestGreen}{$\uparrow$3.03}} \\
GSPO & 47.64{\scriptsize\textcolor{ForestGreen}{$\uparrow$3.59}} & 44.27{\scriptsize\textcolor{ForestGreen}{$\uparrow$4.65}} & 25.20{\scriptsize\textcolor{ForestGreen}{$\uparrow$2.50}} & 66.40{\scriptsize\textcolor{ForestGreen}{$\uparrow$5.30}} & 59.15{\scriptsize\textcolor{ForestGreen}{$\uparrow$1.56}} & 48.53{\scriptsize\textcolor{ForestGreen}{$\uparrow$3.52}} \\
SAPO & 49.46{\scriptsize\textcolor{ForestGreen}{$\uparrow$5.41}} & 44.78{\scriptsize\textcolor{ForestGreen}{$\uparrow$5.16}} & 27.63{\scriptsize\textcolor{ForestGreen}{$\uparrow$4.93}} & 66.00{\scriptsize\textcolor{ForestGreen}{$\uparrow$4.90}} & 61.82{\scriptsize\textcolor{ForestGreen}{$\uparrow$4.23}} & 49.94{\scriptsize\textcolor{ForestGreen}{$\uparrow$4.93}} \\
\midrule
\rowcolor{gray!10} \multicolumn{7}{l}{\textit{Fine-Grained Credit Baselines}} \\
StepGRPO & 41.12{\scriptsize\textcolor{red}{$\downarrow$2.93}} & 35.34{\scriptsize\textcolor{red}{$\downarrow$4.28}} & 19.80{\scriptsize\textcolor{red}{$\downarrow$2.90}} & 59.70{\scriptsize\textcolor{red}{$\downarrow$1.40}} & 55.23{\scriptsize\textcolor{red}{$\downarrow$2.36}} & 42.24{\scriptsize\textcolor{red}{$\downarrow$2.77}} \\
FT-RL & 44.19{\scriptsize\textcolor{ForestGreen}{$\uparrow$0.14}} & 39.14{\scriptsize\textcolor{red}{$\downarrow$0.48}} & 22.37{\scriptsize\textcolor{red}{$\downarrow$0.33}} & 61.20{\scriptsize\textcolor{ForestGreen}{$\uparrow$0.10}} & 57.35{\scriptsize\textcolor{red}{$\downarrow$0.24}} & 44.85{\scriptsize\textcolor{red}{$\downarrow$0.16}} \\
VPPO & 48.89{\scriptsize\textcolor{ForestGreen}{$\uparrow$4.84}} & 45.12{\scriptsize\textcolor{ForestGreen}{$\uparrow$5.50}} & 28.03{\scriptsize\textcolor{ForestGreen}{$\uparrow$5.33}} & 65.80{\scriptsize\textcolor{ForestGreen}{$\uparrow$4.70}} & 60.97{\scriptsize\textcolor{ForestGreen}{$\uparrow$3.38}} & 49.76{\scriptsize\textcolor{ForestGreen}{$\uparrow$4.75}} \\
\midrule
\rowcolor{gray!10} \multicolumn{7}{l}{\textit{Anchor-RL (Ours)}} \\
GRPO + AT-RL & 49.24{\scriptsize\textcolor{ForestGreen}{$\uparrow$5.19}} & 45.98{\scriptsize\textcolor{ForestGreen}{$\uparrow$6.36}} & 27.97{\scriptsize\textcolor{ForestGreen}{$\uparrow$5.27}} & 66.70{\scriptsize\textcolor{ForestGreen}{$\uparrow$5.60}} & 62.73{\scriptsize\textcolor{ForestGreen}{$\uparrow$5.14}} & 50.52{\scriptsize\textcolor{ForestGreen}{$\uparrow$5.51}} \\
DAPO + AT-RL & 50.13{\scriptsize\textcolor{ForestGreen}{$\uparrow$6.08}} & 47.63{\scriptsize\textcolor{ForestGreen}{$\uparrow$8.01}} & 28.73{\scriptsize\textcolor{ForestGreen}{$\uparrow$6.03}} & 68.10{\scriptsize\textcolor{ForestGreen}{$\uparrow$7.00}} & 63.56{\scriptsize\textcolor{ForestGreen}{$\uparrow$5.97}} & 51.63{\scriptsize\textcolor{ForestGreen}{$\uparrow$6.62}} \\
GSPO + AT-RL & 50.74{\scriptsize\textcolor{ForestGreen}{$\uparrow$6.69}} & 47.46{\scriptsize\textcolor{ForestGreen}{$\uparrow$7.84}} & 29.32{\scriptsize\textcolor{ForestGreen}{$\uparrow$6.62}} & 68.10{\scriptsize\textcolor{ForestGreen}{$\uparrow$7.00}} & 64.46{\scriptsize\textcolor{ForestGreen}{$\uparrow$6.87}} & 52.02{\scriptsize\textcolor{ForestGreen}{$\uparrow$7.01}} \\
\rowcolor{blue!5}
\textbf{SAPO + AT-RL} & \textbf{52.77{\scriptsize\textcolor{ForestGreen}{$\uparrow$8.72}}} & \textbf{48.34{\scriptsize\textcolor{ForestGreen}{$\uparrow$8.72}}} & \textbf{30.33{\scriptsize\textcolor{ForestGreen}{$\uparrow$7.63}}} & \textbf{68.60{\scriptsize\textcolor{ForestGreen}{$\uparrow$7.50}}} & \textbf{66.23{\scriptsize\textcolor{ForestGreen}{$\uparrow$8.64}}} & \textbf{53.25{\scriptsize\textcolor{ForestGreen}{$\uparrow$8.24}}} \\
\bottomrule
\end{tabular}
}
\end{table}

% ... [Main Results text unchanged] ...

\subsection{Generalization \& Scalability}
\label{subsec:generalization}

% ... [Cross-Dataset text unchanged] ...

\begin{table}[tb!]
\centering
\caption{
Cross-dataset generalization of AT-RL on Qwen2.5-VL-7B and Qwen2.5-VL-3B.
}
\label{tab:cross_dataset}
\resizebox{\textwidth}{!}{
\begin{tabular}{l|ccccc|c}
\toprule
\textbf{Method} & GeoQA$_{\text{test}}$ & MathVerse & MathVision & MathVista & WeMath & \textbf{Average} \\
\midrule
\multicolumn{7}{c}{\textbf{Qwen2.5-VL-7B}} \\
\midrule
Zero-shot & 44.05 & 39.62 & 22.70 & 61.10 & 57.59 & 45.01 \\
GRPO & 45.92{\scriptsize\textcolor{ForestGreen}{$\uparrow$1.87}} & 41.48{\scriptsize\textcolor{ForestGreen}{$\uparrow$1.86}} & 24.35{\scriptsize\textcolor{ForestGreen}{$\uparrow$1.65}} & 62.75{\scriptsize\textcolor{ForestGreen}{$\uparrow$1.65}} & 59.32{\scriptsize\textcolor{ForestGreen}{$\uparrow$1.73}} & 46.76{\scriptsize\textcolor{ForestGreen}{$\uparrow$1.75}} \\
DAPO & 46.95{\scriptsize\textcolor{ForestGreen}{$\uparrow$2.90}} & 42.75{\scriptsize\textcolor{ForestGreen}{$\uparrow$3.13}} & 25.30{\scriptsize\textcolor{ForestGreen}{$\uparrow$2.60}} & 64.10{\scriptsize\textcolor{ForestGreen}{$\uparrow$3.00}} & 60.75{\scriptsize\textcolor{ForestGreen}{$\uparrow$3.16}} & 47.97{\scriptsize\textcolor{ForestGreen}{$\uparrow$2.96}} \\
GSPO & 46.80{\scriptsize\textcolor{ForestGreen}{$\uparrow$2.75}} & 43.50{\scriptsize\textcolor{ForestGreen}{$\uparrow$3.88}} & 25.15{\scriptsize\textcolor{ForestGreen}{$\uparrow$2.45}} & 65.10{\scriptsize\textcolor{ForestGreen}{$\uparrow$4.00}} & 59.20{\scriptsize\textcolor{ForestGreen}{$\uparrow$1.61}} & 47.95{\scriptsize\textcolor{ForestGreen}{$\uparrow$2.94}} \\
FT-RL & 43.20{\scriptsize\textcolor{red}{$\downarrow$0.85}} & 38.10{\scriptsize\textcolor{red}{$\downarrow$1.52}} & 21.50{\scriptsize\textcolor{red}{$\downarrow$1.20}} & 60.20{\scriptsize\textcolor{red}{$\downarrow$0.90}} & 56.80{\scriptsize\textcolor{red}{$\downarrow$0.79}} & 43.96{\scriptsize\textcolor{red}{$\downarrow$1.05}} \\
\rowcolor{blue!5}
\textbf{DAPO + AT-RL (Ours)} & \textbf{48.55{\scriptsize\textcolor{ForestGreen}{$\uparrow$4.50}}} & \textbf{45.30{\scriptsize\textcolor{ForestGreen}{$\uparrow$5.68}}} & \textbf{27.40{\scriptsize\textcolor{ForestGreen}{$\uparrow$4.70}}} & \textbf{66.50{\scriptsize\textcolor{ForestGreen}{$\uparrow$5.40}}} & \textbf{62.10{\scriptsize\textcolor{ForestGreen}{$\uparrow$4.51}}} & \textbf{49.97{\scriptsize\textcolor{ForestGreen}{$\uparrow$4.96}}} \\
\midrule
\multicolumn{7}{c}{\textbf{Qwen2.5-VL-3B}} \\
\midrule
Zero-shot & 38.72 & 32.97 & 17.96 & 55.20 & 51.78 & 39.33 \\
GRPO & 40.65{\scriptsize\textcolor{ForestGreen}{$\uparrow$1.93}} & 34.92{\scriptsize\textcolor{ForestGreen}{$\uparrow$1.95}} & 19.65{\scriptsize\textcolor{ForestGreen}{$\uparrow$1.69}} & 57.15{\scriptsize\textcolor{ForestGreen}{$\uparrow$1.95}} & 53.62{\scriptsize\textcolor{ForestGreen}{$\uparrow$1.84}} & 41.20{\scriptsize\textcolor{ForestGreen}{$\uparrow$1.87}} \\
DAPO & 41.70{\scriptsize\textcolor{ForestGreen}{$\uparrow$2.98}} & 35.95{\scriptsize\textcolor{ForestGreen}{$\uparrow$2.98}} & 20.60{\scriptsize\textcolor{ForestGreen}{$\uparrow$2.64}} & 58.15{\scriptsize\textcolor{ForestGreen}{$\uparrow$2.95}} & 54.65{\scriptsize\textcolor{ForestGreen}{$\uparrow$2.87}} & 42.21{\scriptsize\textcolor{ForestGreen}{$\uparrow$2.88}} \\
GSPO & 41.65{\scriptsize\textcolor{ForestGreen}{$\uparrow$2.93}} & 36.10{\scriptsize\textcolor{ForestGreen}{$\uparrow$3.13}} & 20.45{\scriptsize\textcolor{ForestGreen}{$\uparrow$2.49}} & 58.30{\scriptsize\textcolor{ForestGreen}{$\uparrow$3.10}} & 54.70{\scriptsize\textcolor{ForestGreen}{$\uparrow$2.92}} & 42.24{\scriptsize\textcolor{ForestGreen}{$\uparrow$2.91}} \\
FT-RL & 36.20{\scriptsize\textcolor{red}{$\downarrow$2.52}} & 30.10{\scriptsize\textcolor{red}{$\downarrow$2.87}} & 15.20{\scriptsize\textcolor{red}{$\downarrow$2.76}} & 52.80{\scriptsize\textcolor{red}{$\downarrow$2.40}} & 49.20{\scriptsize\textcolor{red}{$\downarrow$2.58}} & 36.70{\scriptsize\textcolor{red}{$\downarrow$2.63}} \\
\rowcolor{blue!5}
\textbf{DAPO + AT-RL (Ours)} & \textbf{43.85{\scriptsize\textcolor{ForestGreen}{$\uparrow$5.13}}} & \textbf{38.90{\scriptsize\textcolor{ForestGreen}{$\uparrow$5.93}}} & \textbf{22.80{\scriptsize\textcolor{ForestGreen}{$\uparrow$4.84}}} & \textbf{60.90{\scriptsize\textcolor{ForestGreen}{$\uparrow$5.70}}} & \textbf{58.30{\scriptsize\textcolor{ForestGreen}{$\uparrow$6.52}}} & \textbf{44.95{\scriptsize\textcolor{ForestGreen}{$\uparrow$5.62}}} \\
\bottomrule
\end{tabular}
}
\end{table}

% ... [Model Scale Scaling text unchanged] ...

\begin{table}[ht]
\centering
\caption{Performance (\%) of large multimodal models on diverse benchmarks. All scores are zero-shot unless otherwise noted.}
\label{tab:multimodal-benchmark-vertical}
\resizebox{\textwidth}{!}{
\begin{tabular}{lccccccccc}
\toprule
\textbf{Category} & \textbf{Benchmark} & \textbf{Qwen2.5-VL-32B} & \textbf{Qwen2.5-VL-32B} & \textbf{Qwen2.5-VL-32B} & \textbf{Llama 4} & \textbf{Claude 3.5} & \textbf{Gemini 2.0} & \textbf{OpenAI} & \textbf{Qwen2.5-VL-72B} \\
& & \textit{Instruct} & \textit{+VPPO} & \textit{+Ours} & \textit{Scout} & \textit{Sonnet} & \textit{Flash} & \textit{GPT-4o} & \textit{Instruct} \\
\midrule
\multirow{3}{*}{\textit{Mathematical}}
& MathVista\_testmini & 75.7 & 78.3 & \cellcolor{mycolor} 80.2 & 70.7 & 67.7 & 73.4 & 60.0 & 77.8 \\
& MathVerse\_testmini & 49.3 & 54.5 & \cellcolor{mycolor} 56.6 & -- & 47.8 & 54.6 & 41.2 & 57.2 \\
& MathVision\_test & 38.2 & 42.3 & \cellcolor{mycolor} 44.2 & -- & 33.5 & 41.3 & 30.6 & 43.1 \\
\midrule
\multirow{3}{*}{\textit{Multi-Discipline}}
& MMMU-Pro & 48.5 & 49.2 & \cellcolor{mycolor} 51.9 & 52.2 & 51.5 & 51.7 & 51.9 & 51.6 \\
& MMMU\_Val & 31.7 & 34.1 & \cellcolor{mycolor} 36.5 & -- & 30.5 & 32.9 & 31.1 & 37.2 \\
& EMMA\_core & 14.7 & 17.8 & \cellcolor{mycolor} 19.4 & 24.6 & 18.7 & 17.2 & 6.3 & 17.7 \\
\midrule
\multirow{1}{*}{\textit{Real World}}
& MEGA & 48.9 & 52.3 & \cellcolor{mycolor} 53.0 & 31.8 & 52.5 & 54.7 & 52.7 & 49.6 \\
\midrule
\multirow{2}{*}{\textit{Chart VQA}}
& ChartQAPro & 43.4 & 46.6 & \cellcolor{mycolor} 48.4 & -- & 55.8 & 53.7 & 46.9 & 47.2 \\
& chatmimic & 78.8 & 79.2 & \cellcolor{mycolor} 79.4 & -- & 65.4 & -- & 83.2 & -- \\
\midrule
\multirow{2}{*}{\textit{General VQA}}
& TextVQA & 84.6 & 86.2 & \cellcolor{mycolor} 85.3 & 62.5 & 75.2 & 78.1 & 80.5 & 84.3 \\
& InfoVQA & -- & -- & \cellcolor{mycolor} 87.1 & 60.1 & 72.8 & 75.6 & 77.0 & 76.9 \\
\bottomrule
\end{tabular}
}
\end{table}

\noindent \textbf{Modality Generalization to Video.}  
Table~\ref{tab:video_reasoning} demonstrates AT-RL's extensibility beyond static images to temporal video understanding. Our T-GRPO+AT-RL variant consistently outperforms zero-shot, SFT, and T-GRPO baselines across all frame settings. At 64 frames—the most informative setting—our method achieves \textbf{56.8} average accuracy, a \textbf{+9.4pp} gain over zero-shot and \textbf{+5.1pp} over T-GRPO. Gains are especially pronounced on VSI-Bench (+11.8pp over zero-shot), indicating that AT-RL effectively identifies critical temporal anchor tokens for video reasoning.

\begin{table}[tb!]
\centering
\caption{
Video reasoning performance of \textbf{Qwen2.5-VL-7B} on the Video-R1 benchmark under different frame inputs (16, 32, 64).
Arrow values represent absolute percentage-point changes relative to the \emph{Zero-shot baseline}.
{\scriptsize\textcolor{ForestGreen}{$\uparrow$}} indicates improvement.
The best results are highlighted in \textbf{bold}.
}
\label{tab:video_reasoning}
\resizebox{\textwidth}{!}{
\begin{tabular}{l|ccc|ccc|ccc}
\toprule
\multicolumn{1}{c|}{} & \multicolumn{3}{c|}{\textbf{16 Frames}} & \multicolumn{3}{c|}{\textbf{32 Frames}} & \multicolumn{3}{c}{\textbf{64 Frames}} \\
\cmidrule(lr){2-4} \cmidrule(lr){5-7} \cmidrule(l){8-10}
\textbf{Method} & \textbf{VSI} & \textbf{VidMMMU} & \textbf{MMVU} & 
\textbf{VSI} & \textbf{VidMMMU} & \textbf{MMVU} & 
\textbf{VSI} & \textbf{VidMMMU} & \textbf{MMVU} \\
\midrule
Zero-shot   & 27.7 & 47.8 & 59.2 & 30.1 & 48.1 & 60.0 & 32.5 & 48.5 & 61.2 \\
SFT         & 31.5{\tiny↑3.8} & 49.2{\tiny↑1.4} & 60.8{\tiny↑1.6} & 34.2{\tiny↑4.1} & 50.5{\tiny↑2.4} & 62.3{\tiny↑2.3} & 36.8{\tiny↑4.3} & 51.0{\tiny↑2.5} & 63.5{\tiny↑2.3} \\
T-GRPO      & 33.8{\tiny↑6.1} & 48.9{\tiny↑1.1} & 62.1{\tiny↑2.9} & 36.5{\tiny↑6.4} & 50.2{\tiny↑2.1} & 64.0{\tiny↑4.0} & 39.2{\tiny↑6.7} & 50.8{\tiny↑2.3} & 65.0{\tiny↑3.8} \\
\rowcolor{blue!5}
\textbf{Ours} & \textbf{36.4{\tiny↑8.7}} & \textbf{51.5{\tiny↑3.7}} & \textbf{64.8{\tiny↑5.6}} & 
\textbf{39.6{\tiny↑9.5}} & \textbf{53.4{\tiny↑5.3}} & \textbf{67.2{\tiny↑7.2}} & 
\textbf{44.3{\tiny↑11.8}} & \textbf{56.0{\tiny↑7.5}} & \textbf{70.2{\tiny↑9.0}} \\
\bottomrule
\end{tabular}
}
\end{table}

\subsection{Ablation Studies}
\label{subsec:ablation}

\noindent \textbf{Soft Weighting Necessity.}  
Table~\ref{tab:ablation_hard_truncation} compares hard truncation strategies against our full soft-weighting scheme. While top-15\% hard truncation yields modest gains (+2.96pp), it remains significantly inferior to full AT-RL (+5.65pp). Random or reverse weighting severely degrades performance (-1.52pp / -2.82pp), confirming that indiscriminate token selection introduces harmful noise. This validates our design choice: \textit{soft}, density-proportional weighting preserves gradient signal continuity while emphasizing anchor tokens.

\noindent
\begin{minipage}[t]{0.48\textwidth}
\centering
\captionof{table}{Ablation study on hard truncation strategies based on cross-modal connectivity density.}
\label{tab:ablation_hard_truncation}
\resizebox{\textwidth}{!}{
\begin{tabular}{lc}
\toprule
\textbf{Method} & \textbf{MathVerse} \\
\midrule
Zero-shot & 39.62 \\
GRPO (Uniform weighting) & 41.48{\scriptsize\textcolor{ForestGreen}{$\uparrow$1.86}} \\
Random weighted & 38.10{\scriptsize\textcolor{red}{$\downarrow$1.52}} \\
Reverse weighted & 36.80{\scriptsize\textcolor{red}{$\downarrow$2.82}} \\
Hard truncation (top 5\%) & 41.12{\scriptsize\textcolor{ForestGreen}{$\uparrow$1.50}} \\
Hard truncation (top 10\%) & 42.12{\scriptsize\textcolor{ForestGreen}{$\uparrow$2.50}} \\
Hard truncation (top 15\%) & 42.58{\scriptsize\textcolor{ForestGreen}{$\uparrow$2.96}} \\
Hard truncation (top 20\%) & 42.42{\scriptsize\textcolor{ForestGreen}{$\uparrow$2.80}} \\
Hard truncation (top 30\%) & 41.95{\scriptsize\textcolor{ForestGreen}{$\uparrow$2.33}} \\
Hard truncation (top 50\%) & 41.68{\scriptsize\textcolor{ForestGreen}{$\uparrow$2.06}} \\
Hard truncation (top 75\%) & 41.30{\scriptsize\textcolor{ForestGreen}{$\uparrow$1.68}} \\
\rowcolor{blue!5}
\textbf{AT-RL (Soft weighting)} & \textbf{45.27{\scriptsize\textcolor{ForestGreen}{$\uparrow$5.65}}} \\
\bottomrule
\end{tabular}
}
\vspace{1em}
\end{minipage}
\hfill
\begin{minipage}[t]{0.48\textwidth}
\centering
\captionof{table}{Ablation study on bias correction for AT-RL on MathVerse benchmark.}
\label{tab:ablation_bias_correction}
\resizebox{\textwidth}{!}{
\begin{tabular}{lc}
\toprule
\textbf{Method} & \textbf{MathVerse} \\
\midrule
Zero-shot & 32.97 \\
GRPO (Uniform) & 35.10{\scriptsize\textcolor{ForestGreen}{$\uparrow$2.13}} \\
AT-RL w/o Bias Correction & 34.20{\scriptsize\textcolor{ForestGreen}{$\uparrow$1.23}} \\
\rowcolor{blue!5}
\textbf{AT-RL w/ Bias Correction} & \textbf{37.50{\scriptsize\textcolor{ForestGreen}{$\uparrow$4.53}}} \\
\bottomrule
\end{tabular}
}

\vspace{2em}

\centering
\captionof{table}{Ablation study of semantic grouping mechanisms in AT-RL on MathVerse and MathVista. Standard deviations $(\pm)$ are reported across 3 independent runs.}
\label{tab:ablation_grouping}
\resizebox{\textwidth}{!}{
\begin{tabular}{l|lc|lc}
\toprule
\textbf{Configuration} & \textbf{MathVerse} & $\Delta$ & \textbf{MathVista} & $\Delta$ \\
\midrule
Zero-shot (Base) & 39.62 & - & 61.10 & - \\
Standard GRPO (Uniform Credit) & 41.48 {\scriptsize $\pm$ 0.18} {\scriptsize\textcolor{ForestGreen}{$\uparrow$1.86}} & -3.12 & 62.75 {\scriptsize $\pm$ 0.20} {\scriptsize\textcolor{ForestGreen}{$\uparrow$1.65}} & -2.85 \\
\textit{Impact of Grouping Structure} & & & & \\
-- w/o Grouping (\textbf{Raw Attention Score}) & 42.35 {\scriptsize $\pm$ 0.22} {\scriptsize\textcolor{ForestGreen}{$\uparrow$2.73}} & -2.25 & 63.30 {\scriptsize $\pm$ 0.21} {\scriptsize\textcolor{ForestGreen}{$\uparrow$2.20}} & -2.30 \\
-- Grouping Only (Base METIS) & 43.18 {\scriptsize $\pm$ 0.19} {\scriptsize\textcolor{ForestGreen}{$\uparrow$3.56}} & -1.42 & 64.30 {\scriptsize $\pm$ 0.17} {\scriptsize\textcolor{ForestGreen}{$\uparrow$3.20}} & -1.30 \\
\textit{Impact of Structural Optimization} & & & & \\
-- w/o Neighborhood Expansion & 43.88 {\scriptsize $\pm$ 0.15} {\scriptsize\textcolor{ForestGreen}{$\uparrow$4.26}} & -0.72 & 64.95 {\scriptsize $\pm$ 0.14} {\scriptsize\textcolor{ForestGreen}{$\uparrow$3.85}} & -0.65 \\
-- w/o Cluster Refinement & 43.78 {\scriptsize $\pm$ 0.18} {\scriptsize\textcolor{ForestGreen}{$\uparrow$4.16}} & -0.82 & 64.90 {\scriptsize $\pm$ 0.14} {\scriptsize\textcolor{ForestGreen}{$\uparrow$3.80}} & -0.70 \\
\rowcolor{blue!5}
\textbf{Full AT-RL (Grouping+Refine+Expand)} & \textbf{44.60 {\scriptsize $\pm$ 0.12} {\scriptsize\textcolor{ForestGreen}{$\uparrow$4.98}}} & \textbf{0.00} & \textbf{65.60 {\scriptsize $\pm$ 0.11} {\scriptsize\textcolor{ForestGreen}{$\uparrow$4.50}}} & \textbf{0.00} \\
\bottomrule
\end{tabular}
}
\end{minipage}

\noindent \textbf{Computational Efficiency.} To assess the practical scalability of AT-RL, we profiled its computational footprint on an NVIDIA A100 GPU. Our analysis of a single training iteration reveals that the core training steps—the backward and forward passes—dominate the execution time at \textbf{33.5\%} and \textbf{65.4\%}, respectively. In stark contrast, the entire AT-RL module (including attention processing, bias correction, clustering, and weight computation) introduces a negligible overhead of only \textbf{1.2\%}. This minimal latency confirms that AT-RL functions as a highly efficient, plug-and-play enhancement that facilitates large-scale RLVR training without introducing significant computational bottlenecks.
\section{Discussion}

\begin{figure}[tb!]
    \centering
    \includegraphics[width=1\linewidth]{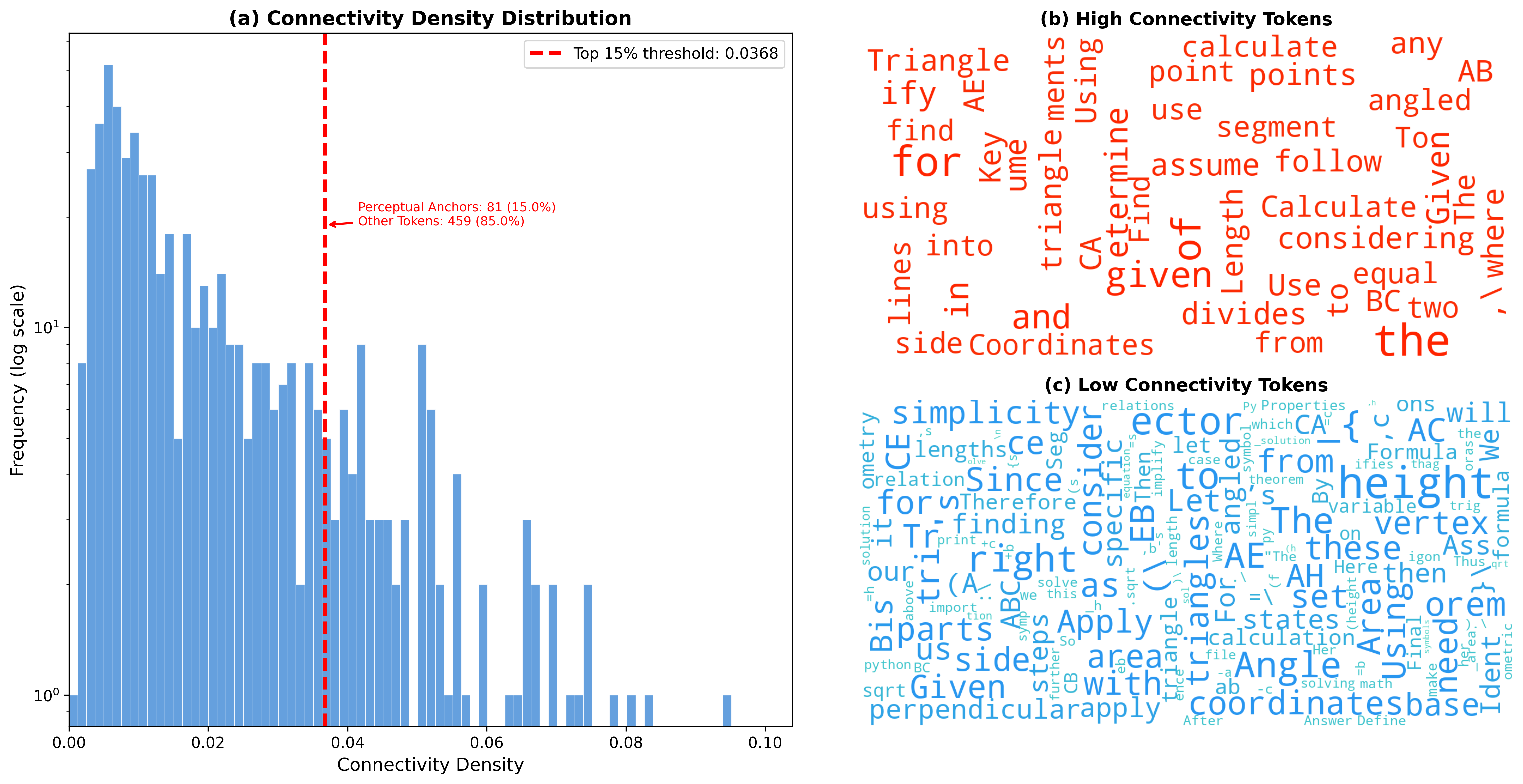}
    \caption{Cross-modal connectivity patterns in multimodal Chain-of-Thought reasoning. (a) Distribution of connectivity density across generated tokens. A minority of tokens exhibit high connectivity, while the majority have low connectivity. (b) \& (c) Word clouds of the top 100 tokens with the highest and lowest average connectivity density, respectively, selected from frequently occurring tokens. A larger font size indicates higher average connectivity. Tokens with high connectivity typically serve as perceptual anchors that ground reasoning in visual evidence, whereas tokens with low connectivity primarily maintain linguistic coherence along the reasoning path.}
    \label{fig:placeholder}
\end{figure}

\textbf{Discussion 1: Perceptual Anchors Emerge in Multimodal CoT.}
We analyze the cross-modal connectivity patterns in multimodal Chain-of-Thought reasoning by computing the average attention weight each generated token receives from visual patches across all layers and heads. As shown in Figure~\ref{fig:placeholder}(a), the connectivity density distribution over 540 tokens from 100 MathVision samples is highly skewed: only 81 tokens (15.0\%) exhibit high connectivity (above the top-15\% threshold of 0.0368), while the remaining 459 tokens (85.0\%) have low connectivity. This bimodal structure suggests a functional dichotomy. The high-connectivity tokens, visualized in Figure~\ref{fig:placeholder}(b), are dominated by action-oriented, reasoning-directing words such as ``calculate'', ``given'', ``determine'', ``assume'', and ``segment'', indicating their role as perceptual anchors that bridge visual perception and logical derivation. In contrast, the low-connectivity tokens in Figure~\ref{fig:placeholder}(c) consist largely of abstract terms like ``simplify'', ``relation'', ``vector'', ``formula'', and syntactic fragments such as ``since'' and ``therefore'', reflecting their function in maintaining linguistic coherence and structural flow. These findings confirm that cross-modal connectivity correlates strongly with semantic function, providing the empirical foundation for our anchor-token reinforcement strategy.

\begin{figure}[tb!]
    \centering
    \includegraphics[width=1\linewidth]{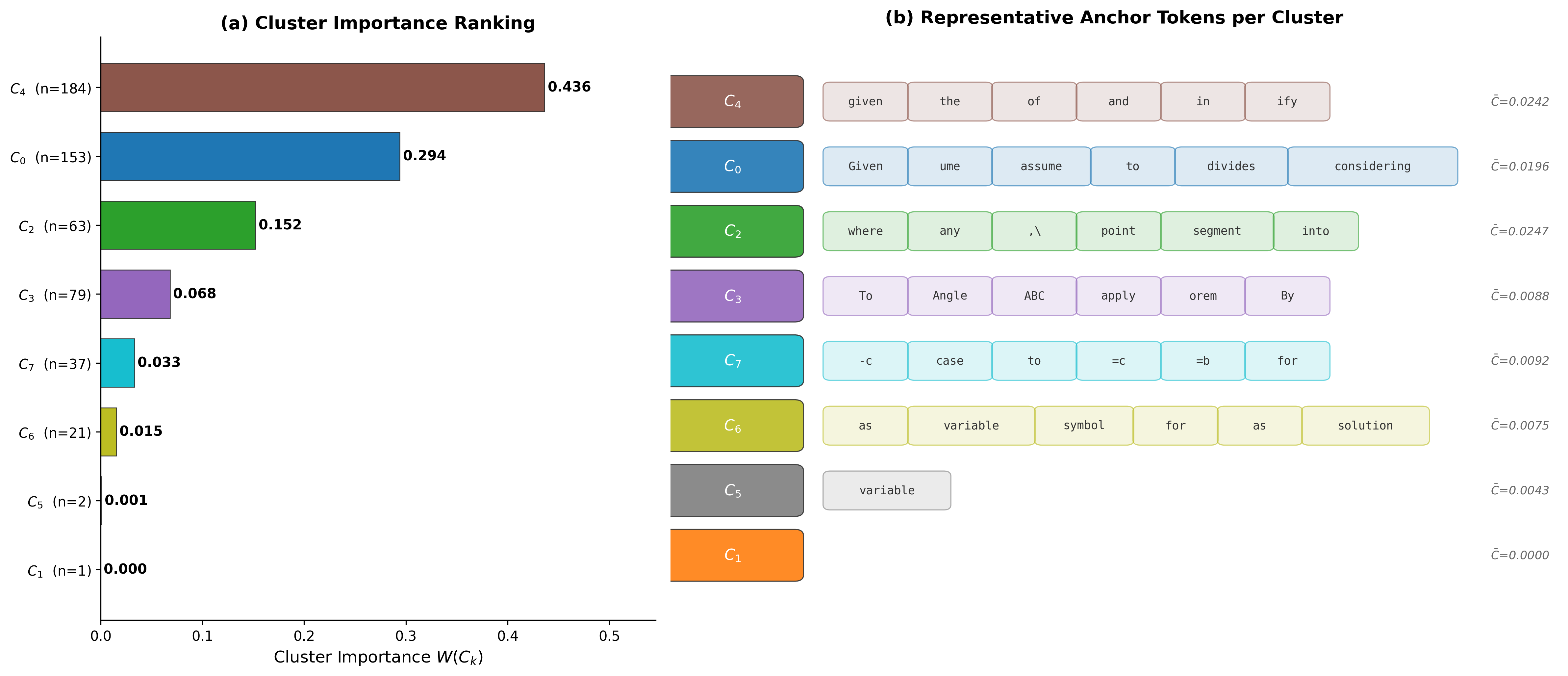}
    \caption{Case study of perceptual anchor clusters in MathVision reasoning. (a) Importance ranking of token clusters based on average connectivity density. (b) Representative tokens for each cluster, illustrating the semantic difference between high-importance and low-importance clusters.}
    \label{fig:anchor_clusters}
    \vspace{-8pt}
\end{figure}

\textbf{Discussion 2: Why Cluster Anchors? A Case Study.}
Assigning credit at the individual token level is noisy due to surface-form variation and positional instability. To identify stable grounding signals, we cluster high-connectivity tokens by semantic and contextual similarity.

Figure~\ref{fig:anchor_clusters}(a) shows that anchor importance is highly concentrated: clusters $C_4$ and $C_0$ dominate the distribution, together accounting for the majority of total importance. This justifies grouping tokens—credit assignment benefits from operating on coherent units rather than isolated instances.

Figure~\ref{fig:anchor_clusters}(b) reveals that these top clusters consist of tokens consistently associated with visual grounding. $C_4$ contains frequent function words such as \textit{given}, \textit{the}, and \textit{of}, which commonly introduce references to diagram elements. $C_0$ includes action-oriented terms like \textit{assume}, \textit{divides}, and \textit{considering}, which trigger interpretation of spatial or geometric relations in the image. The consistency of these patterns across tokens demonstrates that clustering captures reliable grounding behaviors that would be obscured at the single-token level.

\begin{figure}[tb!]
    \centering
    \includegraphics[width=1\linewidth]{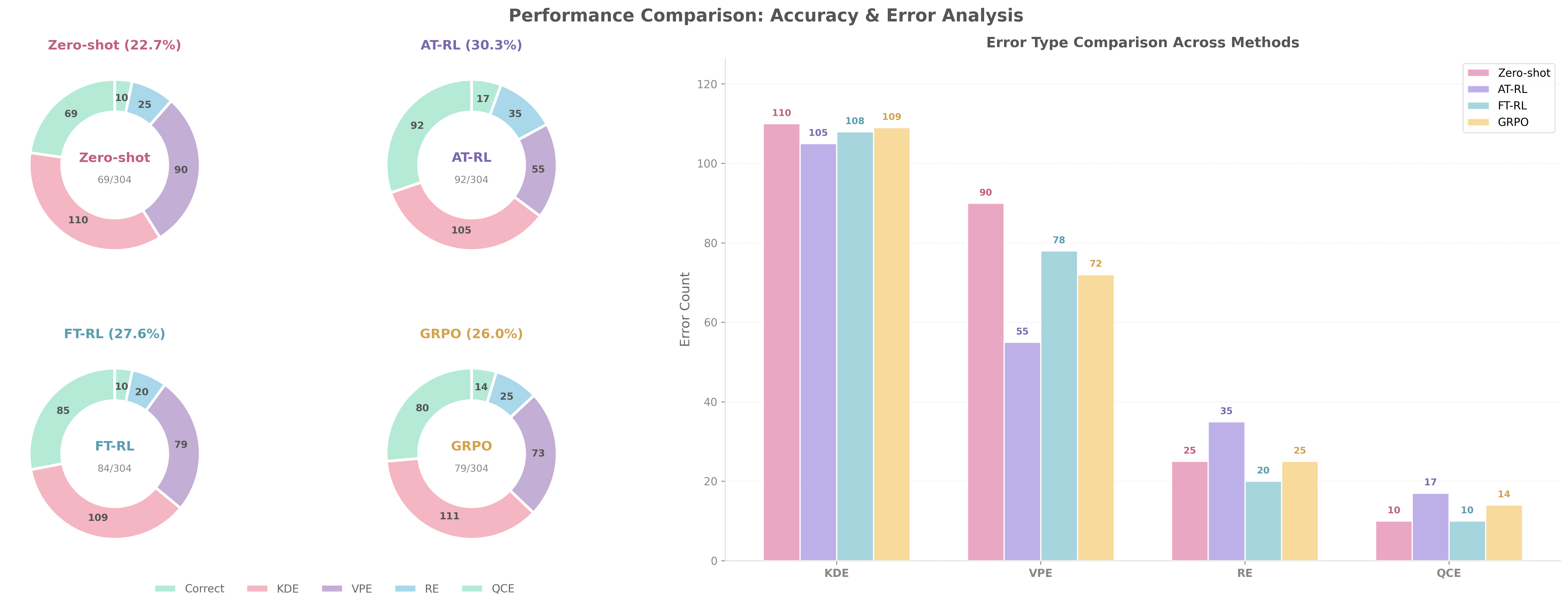}
    \caption{Performance and error breakdown on MathVision (304 questions): accuracy and error composition for Zero-shot (22.7\%), AT-RL (30.3\%), FT-RL (27.6\%), and GRPO (26.0\%). Right: absolute error counts by type.}
    \label{fig:performance_comparison}
    \vspace{-8pt}
\end{figure}

\textbf{Discussion 3: Credit Goes to Anchors --- Whether Right or Wrong.}
AT-RL leverages the verifiable reward signal from RLVR to assign credit based on cross-modal connectivity, regardless of whether the final answer is correct or incorrect. When the model produces a correct answer, the positive reward reinforces high-connectivity tokens (perceptual anchors), strengthening their role in grounding reasoning on visual evidence. Conversely, when the answer is wrong, the negative reward is also concentrated on these same anchors, as they are most likely responsible for misinterpreting the visual input. This mechanism ensures that learning signals are consistently aligned with the fidelity of visual grounding, rather than being diluted across all tokens or biased by linguistic priors. In contrast, uniform credit assignment or SFT cannot distinguish the functional role of individual tokens in multimodal reasoning, leading to less efficient and less reliable updates.

\textbf{Discussion 4: Limitations of RL and Complementarity with SFT.}
Figure~\ref{fig:performance_comparison} highlights a key limitation of reinforcement learning: despite AT-RL’s substantial reduction in Visual Perception Errors (VPE), Knowledge Deployment Errors (KDE) remain largely unaffected across some RL variants. This is inherent to RL—it optimizes the application of existing knowledge but cannot supply missing domain knowledge.

In contrast, supervised fine-tuning (SFT), or even continual pre-training (CPT), excels at injecting factual and procedural knowledge through the direct imitation of expert reasoning traces. Nevertheless, these paradigms offer no inherent mechanism to guarantee that such knowledge is strictly anchored to visual evidence; consequently, models may favor surface-level heuristics over genuine multimodal alignment. These approaches are thus inherently complementary: whereas SFT and CPT equip the model with the necessary 'content to know,' AT-RL dictates 'how to ground it correctly' by reinforcing perceptual anchors. Integrating both is essential to simultaneously resolve knowledge deficits and grounding deficiencies—the twin pillars of failure in complex multimodal reasoning.
\section{Conclusion}
\label{sec:conclusion}

In this work, we introduced \textbf{AT-RL} (\textbf{A}nchor-\textbf{T}oken \textbf{R}einforcement \textbf{L}earning), a plug-and-play framework that addresses the uniform credit assignment limitation in existing RLVR methods by modulating advantage signals proportional to the semantic importance of perceptual anchors. 
By leveraging graph-based token grouping and topological refinement, AT-RL concentrates reinforcement signals on perceptually critical content, significantly improving both early-stage visual grounding and overall reasoning accuracy. Extensive experiments across diverse multimodal benchmarks demonstrate consistent gains over strong baselines, achieving these improvements without architectural modifications or task-specific reward engineering. Ablation studies further validate the contribution of each component.

\clearpage

% \bibliographystyle{unsrt}
% \bibliography{references}

\clearpage

\clearpage

\section*{Supplementary Material}
\label{sec:suppl}

\begin{table}[htbp]
\centering
\caption{Sensitivity analysis of cluster refinement hyperparameters ($\tau_{\mathrm{cen}}$, $\alpha$) on MathVerse and MathVista \textit{testmini} using Qwen2.5-VL-7B-Instruct trained on Geometry-3K. The default setting is $\tau_{\mathrm{cen}}=0.75, \alpha=0.6$.}
\label{tab:appendix_cluster_refinement}
\begin{tabular}{lcc|cc}
\toprule
& \multicolumn{2}{c|}{\textbf{MathVerse}} & \multicolumn{2}{c}{\textbf{MathVista}} \\
\cmidrule(lr){2-3} \cmidrule(lr){4-5}
\textbf{Configuration} & Acc@1 & $\Delta$ & Acc@1 & $\Delta$ \\
\midrule
Default ($\tau_{\mathrm{cen}}=0.75, \alpha=0.6$) & \textbf{48.34} & \textbf{0.00} & \textbf{68.60} & \textbf{0.00} \\
$\tau_{\mathrm{cen}}=0.70, \alpha=0.6$ & 47.92 & -0.42 & 68.10 & -0.50 \\
$\tau_{\mathrm{cen}}=0.80, \alpha=0.6$ & 47.85 & -0.49 & 68.00 & -0.60 \\
$\tau_{\mathrm{cen}}=0.75, \alpha=0.4$ & 47.70 & -0.64 & 67.90 & -0.70 \\
$\tau_{\mathrm{cen}}=0.75, \alpha=0.8$ & 48.05 & -0.29 & 68.20 & -0.40 \\
\bottomrule
\end{tabular}
\end{table}

\begin{table}[htbp]
\centering
\caption{Sensitivity analysis of neighbourhood expansion hyperparameters ($R$, $\tau_{\mathrm{nb}}$, $q$) on MathVerse and MathVista \textit{testmini} using Qwen2.5-VL-7B-Instruct trained on Geometry-3K. The default setting is $R=4, \tau_{\mathrm{nb}}=0.65, q=0.15$.}
\label{tab:appendix_neighbourhood_expansion}
\begin{tabular}{lcc|cc}
\toprule
& \multicolumn{2}{c|}{\textbf{MathVerse}} & \multicolumn{2}{c}{\textbf{MathVista}} \\
\cmidrule(lr){2-3} \cmidrule(lr){4-5}
\textbf{Configuration} & Acc@1 & $\Delta$ & Acc@1 & $\Delta$ \\
\midrule
Default ($R=4, \tau_{\mathrm{nb}}=0.65, q=0.15$) & \textbf{48.34} & \textbf{0.00} & \textbf{68.60} & \textbf{0.00} \\
$R=3, \tau_{\mathrm{nb}}=0.65, q=0.15$ & 48.00 & -0.34 & 68.20 & -0.40 \\
$R=5, \tau_{\mathrm{nb}}=0.65, q=0.15$ & 48.15 & -0.19 & 68.40 & -0.20 \\
$R=4, \tau_{\mathrm{nb}}=0.60, q=0.15$ & 47.90 & -0.44 & 68.10 & -0.50 \\
$R=4, \tau_{\mathrm{nb}}=0.70, q=0.15$ & 48.10 & -0.24 & 68.30 & -0.30 \\
$R=4, \tau_{\mathrm{nb}}=0.65, q=0.10$ & 48.05 & -0.29 & 68.25 & -0.35 \\
$R=4, \tau_{\mathrm{nb}}=0.65, q=0.20$ & 48.12 & -0.22 & 68.35 & -0.25 \\
\bottomrule
\end{tabular}
\end{table}

\section{Implementation Details}
\label{sec:impl}
This appendix contains engineering details, extended experiments, and formulae omitted from the main text for brevity. All components described here are necessary for full reproduction of \textbf{AT-RL} (\textbf{A}nchor-\textbf{T}oken \textbf{R}einforcement \textbf{L}earning).

\subsection{Bias Correction Details}
The bias curve $b \in \mathbb{R}^T$ is parameterised as:
\begin{equation}
\begin{aligned}
  b_j &= 1 + \lambda_{\exp} e^{-p_j \gamma} + \lambda_{\cos} \cos(\pi p_j), \\
  p_j &= \frac{j}{T}, \quad j=1,\dots,T,
\end{aligned}
\end{equation}
where $\lambda_{\exp}$ controls exponential term weight, $\gamma$ decay rate, and $\lambda_{\cos}$ cosine amplitude.  
Mean-normalise $b$ so $\frac{1}{T}\sum_{j=1}^T b_j = 1$, then divide $\bar{\mathbf{A}}$ element-wise to obtain $\bar{\mathbf{A}}^\beta$.

\noindent\textbf{Default:} $\lambda_{\exp} = 0.15$, $\gamma = 4.0$, $\lambda_{\cos} = 0.05$.

\subsection{Cluster Refinement}
Given an initial METIS clustering $\{C_k\}$:
\begin{equation}
  \mathbf{c}_k = \frac{1}{|C_k|} \sum_{t \in C_k} \bar{\mathbf{A}}^{\beta}_t
\end{equation}
Tokens with $\cos(\bar{\mathbf{A}}^{\beta}_t, \mathbf{c}_k) < \tau_{\mathrm{cen}}$ are down-weighted:
\begin{equation}
  \phi_t \leftarrow
  \begin{cases}
    \phi_t, & \cos(\bar{\mathbf{A}}^{\beta}_t, \mathbf{c}_k) \ge \tau_{\mathrm{cen}} \\
    \alpha \cdot \phi_t, & \text{otherwise}
  \end{cases}
\end{equation}
\noindent\textbf{Default:} $\tau_{\mathrm{cen}}=0.75$, $\alpha=0.6$.

\subsection{Neighbourhood Expansion}
Top-$q$ percentile tokens (by $\phi_t$) promote $R$ nearest neighbours if:
\begin{equation}
\begin{aligned}
  &\lambda_{\mathrm{sim}}\cos(\bar{\mathbf{A}}^{\beta}_{t'}, \bar{\mathbf{A}}^{\beta}_t) + \lambda_{\mathrm{imp}}\phi_{t'} > \tau_{\mathrm{nb}}
\end{aligned}
\end{equation}
\noindent\textbf{Default:} $q=0.15$, $R=4$, $\lambda_{\mathrm{sim}}=0.5$, $\lambda_{\mathrm{imp}}=0.5$, $\tau_{\mathrm{nb}}=0.65$.

\subsection{Hyperparameters for AT-RL Optimisation}
Batch size $512$, Group size $G=8$, group-wise mean/std advantage normalisation, PPO clip $\epsilon=0.2$, KL coef $\beta=0.02$, AdamW lr $1\times 10^{-6}$, wd $0.01$, 1 update/rollout.

\subsection{Notation}
$c_t^{(i)}$: token credit from $W(C_k)$;  
$A^{(i)}$: group-relative sequence advantage;  
$A_{\mathrm{AT}}^{(i,t)}$: token-level advantage;  
$r_t(\theta)$: token-level probability ratio;  
$\tau_{\mathrm{sim}}$: graph edge threshold;  
$\phi_t$: graph importance score.

\subsection{Graph Construction Hyperparameters}
$\tau_{\mathrm{sim}}=0.7$, $K=\max(2,\lfloor T/10\rfloor)$.

\section{Applicability and Plug-and-Play Nature}
\label{sec:applicability}

\subsection{KL-Independence and Flexibility}

\textbf{AT-RL is fundamentally a plug-and-play token-level advantage re-weighting mechanism that is completely agnostic to the choice of regularization strategy.} While our default implementation includes a KL regulariser ($\beta=0.02$) to stabilise policy updates, this is purely optional and not intrinsic to the AT-RL methodology.

The core innovation of AT-RL lies in its anchor-token credit assignment, which operates independently of any divergence control mechanism. This design philosophy enables seamless integration into diverse RL paradigms:

\begin{itemize}
    \item \textbf{KL-Free Optimization}: Set $\beta=0$ to completely remove KL regularization, as demonstrated in our ablation studies (Section~\ref{sec:kl_free_exp}).
    \item \textbf{Alternative Regularizers}: Replace KL with entropy bonuses or trust-region constraints without modifying AT-RL's core logic.
    \item \textbf{Cross-Framework Compatibility}: Integrate into REINFORCE, actor-critic, DPO, or KL-free methods like DAPO by simply substituting the advantage term with $A_{\mathrm{AT}}^{(i,t)}$.
\end{itemize}

\textbf{Key Insight}: AT-RL's token-level credit assignment $c_t^{(i)}$ derived from semantic clustering is orthogonal to the policy update rule. Any RL algorithm using per-token advantages can directly benefit from AT-RL's refined credit signal.

\subsection{KL-Free AT-RL Experiments}
\label{sec:kl_free_exp}

To validate AT-RL's independence from KL regularization, we trained AT-RL(+GRPO) with $\beta=0$:

\begin{table}[h]
\centering
\caption{Acc@1(\%) comparison between KL-regularised and KL-free AT-RL(+GRPO). Performance remains comparable, confirming that AT-RL's effectiveness stems from semantic credit assignment.}
\label{tab:kl_free_results}
\begin{tabular}{lcc}
\toprule
Method & GeoQA$_{\text{test}}$ & MathVerse \\
\midrule
AT-RL ($\beta=0.02$) & 45.41 & 40.68 \\
AT-RL ($\beta=0$)    & 45.27 & 40.55 \\
\bottomrule
\end{tabular}
\end{table}

The negligible performance difference ($<0.2\%$) demonstrates that AT-RL's gains are primarily attributable to its anchor-grounded credit assignment mechanism.

\subsection{Integration with KL-Free Methods}

AT-RL can be directly integrated into recent KL-free optimization methods:

\paragraph{DAPO (Divergence-Agnostic Policy Optimization)} To integrate AT-RL:
\begin{equation}
\begin{aligned}
  \mathcal{L}_{\text{DAPO+AT-RL}} = \mathbb{E}_{t}\bigg[&\min\big(r_t(\theta) A_{\mathrm{AT}}^{(i,t)}, \text{clip}(r_t(\theta), 1-\epsilon_t, 1+\epsilon_t) A_{\mathrm{AT}}^{(i,t)}\big)\bigg]
\end{aligned}
\end{equation}
where $A_{\mathrm{AT}}^{(i,t)} = c_t^{(i)} \cdot A^{(i)}$ is the refined token-level advantage.

\section{AT-RL in Other RL Frameworks}
\label{sec:other_rl}

\subsection{REINFORCE}
Replace the standard advantage $\hat{A}_t$ with AT-RL's token-level advantage:
\begin{equation}
  \nabla_\theta J(\theta) = \mathbb{E}\bigg[\sum_{t=1}^T A_{\mathrm{AT}}^{(i,t)} \cdot \nabla_\theta \log \pi_\theta(a_t | s_t)\bigg]
\end{equation}

\subsection{Actor-Critic}
Substitute the advantage term in the policy loss:
\begin{equation}
  \mathcal{L}_{\text{policy}} = -\mathbb{E}\left[\sum_{t=1}^T A_{\mathrm{AT}}^{(i,t)} \log \pi_\theta(a_t | s_t)\right]
\end{equation}

\subsection{DPO and Preference Optimization}
Weight per-token log-likelihood differences by AT-RL credits:
\begin{equation}
  \mathcal{L}_{\text{DPO+AT-RL}} = -\mathbb{E}\bigg[\log \sigma\bigg(\beta \sum_{t=1}^T c_t^{(i)} \cdot \left(\log \frac{\pi_\theta(y_w^t)}{\pi_{\text{ref}}(y_w^t)} - \log \frac{\pi_\theta(y_l^t)}{\pi_{\text{ref}}(y_l^t)}\right)\bigg)\bigg]
\end{equation}

\section{Attention Extraction \& Modality Labelling}
\label{sec:attention}

We extract attention weights from the final $L=4$ decoder layers, averaging across $H=32$ heads. Visual tokens are assigned as $\mathsf{v}$ (vision) and text tokens as $\mathsf{l}$ (language).

\section{Environment and Complexity}
\label{sec:env}

\textbf{Software Stack}: PyTorch 2.2, CUDA 12.1, PyMetis 2023.
\textbf{Hardware}: 8$\times$ NVIDIA A100-80GB GPUs.

\textbf{Computational Breakdown per Training Iteration}:
\begin{itemize}
    \item Forward pass: 65.4\%
    \item Backward pass: 33.5\%
    \item AT-RL credit computation: 1.2\%
\end{itemize}
The negligible overhead ($<2\%$) confirms AT-RL's efficiency.

\section{Hyperparameter Search}
\label{sec:hyperparam}

We performed grid search over the following:
$\tau_{\mathrm{cen}}\in\{0.70,0.75,0.80\}$, $\alpha\in\{0.4,0.6,0.8\}$, $R\in\{3,4,5\}$, $\tau_{\mathrm{nb}}\in\{0.60,0.65,0.70\}$, $q\in\{0.10,0.15,0.20\}$.

\section{Evaluation Protocol}
\label{app:evaluation}

\subsection{Answer Extraction Strategy}
\begin{itemize}
    \item \textbf{Primary}: Extract answers enclosed in \verb|\boxed{}|.
    \item \textbf{Fallback}: Full-text search for option letters or numerical values.
\end{itemize}

\subsection{Semantic Verification Module}
An LLM-based judge (Qwen2.5-72B-Instruct) evaluates answer correctness:
\begin{itemize}
    \item \textbf{Matching}: Unit-agnostic for numerical answers.
    \item \textbf{Temp}: 0.1; \textbf{Format}: Binary (``Correct''/``Incorrect'').
\end{itemize}

\subsection{Sampling Configuration}
Zero-shot with three randomized prompt templates; Temp 0.1; Max Tokens 4096. $N=8$ independent repetitions for Mean@8.

\subsection{Mean@N Metric}
\begin{equation}
\text{Mean@N} = \frac{\text{Number of Passed Problems}}{\text{Total Problems}}
\end{equation}

\end{document}